\documentclass[10pt,letterpaper]{article}

\usepackage{float}
\usepackage{cogsci}
\usepackage{pslatex}
\usepackage{apacite}

\usepackage{latexsym,amssymb,lastpage}
\usepackage{graphicx,amsfonts}
\usepackage{times,mathptmx,bm,amsmath}

\newenvironment{pf}[1][Proof]{\noindent\textit{#1. } }{\hfill$\square$}
  \newtheorem{thm}{Theorem}

\title{Which Learning Algorithms Can Generalize \\ Identity-Based Rules to Novel Inputs?}
 
\author{{\large \bf Paul Tupper (pft3@sfu.ca)} \\
  Department of Mathematics, Simon Fraser University \\
  Burnaby, BC, V5A 1S6, Canada
  \AND {\large \bf Bobak Shahriari (bshahr@cs.ubc.ca)} \\
  Department of Computer Science, University of British Columbia \\
  Vancouver, BC, V6T 1Z4, Canada.
  }

\begin{document}

\maketitle

\begin{abstract}

We propose a novel framework for the analysis of learning algorithms that allows us to say when such algorithms can and cannot generalize certain patterns from training data to test data. In particular we focus on situations where the rule that must be learned concerns two components of a stimulus being identical. We call such a basis for discrimination an {\emph{identity-based}} rule. Identity-based rules have proven to be  difficult or impossible  for certain types of learning algorithms to acquire from limited datasets. This is in contrast to human behaviour on similar tasks. Here we provide a framework for rigorously establishing which learning algorithms will fail at generalizing identity-based rules to novel stimuli. 
We use this framework to show that such algorithms are unable to generalize identity-based rules to novel inputs unless trained on virtually all possible inputs. We demonstrate these results computationally with a multilayer feedforward neural network.

\end{abstract}

\section{Introduction}

Suppose a subject is asked to learn an artificial language in which all words consist of two letters. 
They are  told that CC, AA, HH, EE, and RR are all examples of valid words in the language but that   GA, EH, RA, ER, MG are not valid words.
Now suppose that the learner is asked whether YY and YZ could be  valid words in the language. Presumably they will say that YY could be a valid word in the language whereas YZ could not be. The obvious feature that all the valid words have in common is that they consist of two identical letters. This feature is not shared by the invalid words.  
We say in this case that the learners have learned an identity-based rule, and are able to generalize the rule to novel inputs.

We do not know if this exact experiment has ever been performed, but there  have been analogous tests in the phonological domain \cite{berent2002, gallagher2013}. In artificial language learning tasks, human subjects are sensitive to identity relations between segments, and are able to generalize them to novel inputs. This kind of effect is not specific to language though: consider a task where subjects are presented with pictures of pairs of socks, and are asked to say whether they form a matching pair.

Surprisingly, given how obvious the above pattern is to human learners, many computer models of  learning are not able to learn identity-based rules like those implicit in the data above, without being presented with nearly all possible inputs. These computational  learners may give the same rating to both the forms YY and YZ, since neither of them have any similarity to the training words in a manner that is deemed relevant by the algorithms. Important classes of such algorithms include basic connectionist algorithms \cite{PDP} and the ``Plain" (Baseline version) of the UCLA Phonotactic Learner \cite{hayeswilson2008}.
There are ways to modify these algorithms to perform better on such tasks, for example, by introducing copying \cite{colavin2010}, special representations of identical segments in the input \cite{gallagher2013}, or weight sharing across connections as is done in convolutional neural networks \cite{lecun1995}.

There are many informal arguments given for why the basic versions of these algorithms cannot learn identity-based rules. Such algorithms are unable to generalize ``outside the training space" \cite{marcusbook}, or ``do not instantiate variables" \cite{berentbook}. Though these terms describe a genuine limitation of such algorithms, they suffer the drawback of not being defined formally. Even though computational learners themselves are clearly defined, whether a particular algorithm is able to learn identity relations or instantiate variables is impossible to determine precisely since the criterion for these conditions is not formalized. Our present goal is to provide a rigorous framework for these informal statements about algorithms, and to provide criteria for when an algorithm cannot generalize identity-based rules to novel inputs.

%
%
%

In the following we define learning \emph{algorithms},  \emph{symmetries} of sets of words, and what it means for an algorithm to be \emph{invariant} under a symmetry. In our main result we show that if an algorithm is invariant under some symmetry, and the training data is invariant under the same symmetry, then the algorithm cannot learn a grammar that is not invariant under that symmetry.
 As an application, we demonstrate a symmetry that identity-based rules are not invariant under, and then show that a wide class of algorithms are invariant under it. This means that such algorithms cannot learn identity-based grammars with invariant training data, in contrast to human performance on analogous tasks.  We then demonstrate how feed-forward neural networks suffer from these limitations, independent of the number of hidden layers in the network.


\section{Formal Definitions}

We consider a set $W$, which we call the set of words, containing all well-formed inputs.
We stress that in the linguistic case $W$ consists of both words that are good (grammatical) and words that are bad (ungrammatical). In what follows we will consider words to be strings of letters, but individual words can be anything, such as strings of segments or feature bundles.

To fix ideas, in what follows we will often consider a particular example of a set of words: we let $\widetilde{W}$ be the set of all two letter words, where the letters are capitals taken from the English alphabet, such as AA or MG.

We define the training data $D$ to be a collection of word-rating pairs $\langle w, r \rangle$ where $w$ is a word in $W$ and $r$ is a number interpreted as a {\emph{rating}} of how ``good" $w$ is. For example, using the word set $\widetilde{W}$, a dataset $D$ might consist of
\begin{equation} \label{eqn:sampledata}
\langle CC, 1 \rangle, \ \ \langle AA, 1 \rangle, \ \ \langle EE, 1 \rangle, \ \ 
\langle GA, 0 \rangle, \ \ \langle EH, 0 \rangle, \ \ \langle RA, 0 \rangle.  
\end{equation}
This dataset says that CC, AA, and EE have rating 1 (and thus are good words) and that GA, EH, and RA have rating 0 (and thus are bad words).
Alternatively, in a training task where only good words are given to the learner, $D$ might consist only of good words paired with the rating 1.
But there are other possibilities:  words could be paired with  a rating given by their prevalence  in a corpus, for example.

To formally define a learning algorithm, consider what a learning algorithm such as the UCLA Phonotactic Learner \cite{hayeswilson2008} does.  First, a collection of data $D$ is input to the algorithm and used to choose a set of  parameters $p$ in a model. We can formalize this as $p = \mathcal{A}(D)$.  Once we have $p$, given any new input $w$ the algorithm outputs a  {\emph{score}}, which we can formalize as $f(p,w)$. 
 Typically, the computation of $p$ from $D$ is computationally intensive whereas once we have $p$, the score $f(p,w)$ is cheap to evaluate. This matches our experience of human behaviour where learning a language occurs over long periods of time, whereas  judgements of the well-formedness of novel words are readily produced by adult speakers.

 Here we will abstract away issues of parameter setting and computational effort and just view an algorithm as a map that takes a set of training data $D$ and an input $w$ and outputs a rating. We consider the map $\mathcal{L}$ given by
 \[
 \mathcal{L}(D,w) = f(\mathcal{A}(D),w).
  \]
Specifically, a learning algorithm $\mathcal{L}$ is a map that takes training data $D$ and word $w$ and outputs a score $\mathcal{L}(D,w)$. The interpretation is that this score is what you would get if you used the data $D$ to train the algorithm and then used the resulting computational model to evaluate the word $w$.

We note that we interpret both the ratings coupled with words in the training data $D$ and the scores output by the algorithm $\mathcal{L}$ as measures of the goodness of a word. This is natural, since we expect the algorithm to give good scores to words that have high ratings in the training data. However, ratings and scores are distinct in general; for example, ratings in $D$ could be how common a word is in a corpus and scores from $\mathcal{L}$ could be intended to model how well-formed a word is on a scale from 0 to 1.

We define a  symmetry $\sigma$ to be a bijective map from the set of words $W$ to itself; in other words, a map such that $\sigma(w)$ is in $W$ for all $w$ in $W$, and for all $v$ in $W$ there is an $u$ in $W$ such that $\sigma(u)=v$.  
As an example of a symmetry, let $\tilde{\sigma}$ be the map from $\widetilde{W}$ to itself given by 
\begin{equation} \label{eqn:sigmatilde}
\tilde{\sigma}(\mathbb{XY}) = \mathbb{YX},
\end{equation}
for any letters $\mathbb{X}$, $\mathbb{Y}$. Thus the symmetry $\tilde{\sigma}$ reverses the order of letters in two-letter words.

We introduce symmetries in order to analyze algorithms: we are not claiming that they have any psychological or linguistic reality. Indeed, as far as we know all maps that are naturally occurring phonological processes are not symmetries. For one thing, most phonological maps satisfy $\sigma(\sigma(x))=\sigma(x)$ for all $x$ (also known as idempotency \cite{magri}). But this can only happen with a symmetry if $\sigma(x)=x$ for all $x$, meaning that $\sigma$ does nothing.

A word $w$ is invariant under a symmetry $\sigma$ if $\sigma(w)=w$. To apply a symmetry to a set of training data, we say that $\sigma$ just acts on each word in every word-rating pair in the data set, but does not change the rating of that word. So if the word-rating pair $\langle w, r \rangle$ is in $D$, then the pair $\langle \sigma(w), r\rangle$ is in $\sigma(D)$.  For example, if we applied $\tilde{\sigma}$ (as defined in \eqref{eqn:sigmatilde}) to the dataset in \eqref{eqn:sampledata} we would get the dataset
\[
\langle CC, 1 \rangle, \ \ \langle AA, 1 \rangle, \ \ \langle EE, 1 \rangle, \ \ 
\langle AG, 0 \rangle, \ \ \langle HE, 0 \rangle, \ \ \langle AR, 0 \rangle.  
\]

We say that a dataset $D$ is invariant under a symmetry $\sigma$ if $\sigma(D)$ has precisely the same word-rating pairs as $D$. 
The simplest way for data $D$ to be invariant under a symmetry $\sigma$ is if each word in each word-rating pair in $D$ is invariant under $\sigma$. But there are other ways. For example, the symmetry $\tilde{\sigma}$ leaves the data 
\[
\langle BB, 1 \rangle, \ \ \ \langle GG, 2 \rangle, \ \ \  \langle EE, 0 \rangle
\]
invariant  because the words $BB, GG, EE$  are all invariant under $\tilde{\sigma}$. On the other hand
the data
\[
\langle BG, 1 \rangle, \ \ \  \langle GB, 1 \rangle, \ \ \ \langle EA, 2  \rangle, \ \ \ 
\langle AE, 2 \rangle
\]
is also invariant under $\tilde{\sigma}$, but in this case the individual words are not invariant, it is just that $w$ and $\sigma(w)$ always have the same rating in this data set.

We say an algorithm $\mathcal{L}$ is invariant under $\sigma$ if $\mathcal{L}(\sigma( D), \sigma(w))= \mathcal{L}(D,w)$ for all $D$ and $w$.  In words, the rating that the algorithm gives to $w$ when trained on $D$ is the same that the algorithm gives to $\sigma (w)$ when trained on $\sigma (D)$. 

Our main result is a simple consequence  of these definitions.

\begin{thm} \label{thm:mainResult} If algorithm $\mathcal{L}$ and training data $D$ are invariant under symmetry $\sigma$ then 
\[
\mathcal{L}(D,w)= \mathcal{L}(D, \sigma (w)),
\] 
for all $w$ in $W$.
In other words, the algorithm $\mathcal{L}$ gives the same rating to $w$ and $\sigma( w)$ when trained on $D$.
\end{thm}
\begin{pf}
We have 
\[
\mathcal{L}(D,w) = \mathcal{L}(\sigma (D),\sigma (w)) = \mathcal{L}(D, \sigma (w))
\]
where the first equality follows from the invariance of $\mathcal{L}$ under $\sigma$, and the second inequality follows from the invariance of $D$ under $\sigma$.
\end{pf}

{\bf  Example:} 
Consider a language containing 10 letters, each letter having a sonority value between 1 and 5 according to the following table.
(Sonority is an abstract phonological variable, roughly corresponding to how close a segment is to a vowel.)
\begin{table}[!ht]
\begin{center} 
\caption{Segments  in a Hypothetical Language} 
\label{sample-table} 
\vskip 0.12in
\begin{tabular}{ll} 
\hline
segments   &  sonority \\
\hline
A \ \ O        &   5 \\
W \ \ Y       &   4 \\
M \ \ N       &   3 \\
V \ \ Z           &   2 \\
B \ \ D         &  1\\
\hline
\end{tabular} 
\end{center} 
\end{table}
  Words in the language consist of only two letters. Suppose that all words in the language have increasing or constant sonority.
So, BA, MO, ZW, BD could all be words in the language, but AD, AN, and  WV could not be. Consider the letter reversing symmetry $\tilde{\sigma}$ given in \eqref{eqn:sigmatilde}. If you apply $\tilde{\sigma}$ to an ungrammatical word (e.g. AB) you get  a grammatical word (BA). If you apply $\tilde{\sigma}$ to a grammatical word with increasing sonority you get an ungrammatical word. Words with two letters of the same sonority give you back another word with letters of the same sonority.

Now suppose you have a learning algorithm $\mathcal{L}$ that is invariant under $\tilde{\sigma}$.  This means that if you take a data set $D$, train the algorithm on it, and then use the algorithm to evaluate word $w$, you will get the same result if you train the algorithm on $\tilde{\sigma}(D)$ (in which all the words are reversed) and then use the algorithm to evaluate $\tilde{\sigma}(w)$, which is just the reversal of $w$.

Suppose we give the algorithm data $D$ that is invariant under $\tilde{\sigma}$. 
For simplicity we assume that  $D$ consists only of grammatical words each assigned the rating $1$.
In this case, the only way $D$ can be invariant under $\tilde{\sigma}$  is if all the words in $D$ have constant sonority, and for every such word $\mathbb{X} \mathbb{Y}$ in $D$, $\mathbb{Y}\mathbb{X}$ is also in $D$.  Can the algorithm correctly learn the generalization that words in the language must have increasing or level sonority from this data set?  

Theorem~\ref{thm:mainResult} shows that it cannot, as follows. According to the theorem, $\mathcal{L}(D,w) = \mathcal{L}(D, \tilde{\sigma}(w))$. All we need to do is let $w$ equal a word of increasing sonority, such as $BA$, to see that the algorithm with training data $D$ gives the same score to $BA$ and $AB$. Since the first is grammatical and the second is ungrammatical, the algorithm clearly has not learned the correct rule governing grammaticality in the language.
 This is pretty commonsensical: one way to think of it is that there is nothing in the algorithm or the training data to make the algorithm prefer $AB$ to $BA$, since both the algorithm and the training data are invariant under $\tilde{\sigma}$, and $BA = \tilde{\sigma}(AB)$. Of course, this is not necessarily a defect of the algorithm $\mathcal{L}$; if some words with increasing or decreasing sonority were included in $D$, then $D$ would not be invariant under $\tilde{\sigma}$, and $\mathcal{L}$ could learn the grammar.


In the next section we will give a less straightforward example, allowing us to formalize the idea of identity-based rules for learning algorithms.

\section{Identity-Based Rules}

We now use the above result to show that certain algorithms cannot learn identity-based rules unless trained on words containing virtually all letters in the alphabet. That is, the algorithm cannot extend the identity-based rules to words containing letters that it has not explicitly been trained on.
This is in sharp contrast to human learners who are able to generalize identify-based rules (in the phonological context, for example) to segments they have not encountered before\cite{berent2002}.

We return to the example at the beginning of the paper: $\widetilde{W}$ is the set of all words consisting of two letters. We stipulate that grammatical words are those consisting of two identical letters and all other words are ungrammatical. 
Suppose we want the algorithm to learn this grammar, but train it on data omitting any words containing the letters $Y$ and $Z$.
What algorithms will not be able to learn the correct grammar under these conditions?


Define the symmetry $\sigma$ of $W$ by the following:
\[
\sigma( \mathbb{X}_1 Y) = \mathbb{X}_1 Z, \ \ \ 
\sigma( \mathbb{X}_1 Z ) = \mathbb{X}_1 Y, \ \ \ 
\sigma( \mathbb{X}_1 \mathbb{X}_2 ) = \mathbb{X}_1 \mathbb{X}_2,
\]
for all letters $\mathbb{X}_1, \mathbb{X}_2$, with $\mathbb{X}_2$ not equal to $Y$ or $Z$.
In other words, if the second segment is $Y$, $\sigma$ changes it to $Z$, if the second segment is $Z$, $\sigma$ changes it to $Y$, and if the second segment is neither, then the word is unchanged.

Now suppose our training data $D$ contains no words with either segments $Y$ or $Z$ as the second segment.
$D$ may contain both grammatical words (e.g. $CC$) with rating 1 and ungrammatical words (e.g. $CE$) with rating $0$.
 Then $D$ is invariant under $\sigma$. Theorem~\ref{thm:mainResult} shows that if the algorithm $\mathcal{L}$ is also invariant under $\sigma$ then it will give the same rating to $w$ and $\sigma (w)$ for any word $W$ when trained on $D$. In that case we would have that it gives the same rating to the words $YY$ and $YZ$, showing that it cannot learn the identity based grammar.

Below we provide an example of an algorithm invariant under this symmetry. But in general we informally argue that any algorithm that does not in some way explicitly check for identity between the letters, or somehow enforce a similar treatment of those two letters in processing, cannot correctly learn that $YY$ is a more well formed word than $YZ$, if it is never given words with a second letter $Y$ or $Z$ as training data.

\section{Randomized Algorithms}

Many algorithms for learning use randomness at some point in their operation. It may either be in the computation that takes the input data to the parameters $p$ (for example, by which order the input words are used) or in the map from the parameters and a new input word to a word score $s$. In the former case $p=\mathcal{A}(D)$ is a random function of $D$; in the latter $s=f(p,w)$ is a random function of $p$ and $w$. In either case, this leads to $\mathcal{L}(D,w)$ being random  for any fixed $D$ and $w$.

Under these conditions, it is unlikely that invariance of the form described above will hold. Instead we now define invariance of $\mathcal{L}$ under $\sigma$ to be 
\[
\mathbb{E} \mathcal{L} (\sigma (D),\sigma (w)) = \mathbb{E} \mathcal{L} (D,w),
\]
where $\mathbb{E}$ denotes expectation. (If $X$ is a random variable, $\mathbb{E} X$ is approximately what we would get if we took the average of a large number of samples of $X$.)

We now get the same result as before. This is a strictly stronger result than Theorem~\ref{thm:mainResult}, since a deterministic algorithm is just a special case of a randomized algorithm.

\begin{thm} \label{thm:randomResult} If random algorithm $\mathcal{L}$ and training data $D$ are invariant under symmetry $\sigma$ then 
\[
\mathbb{E} \mathcal{L}(D,w)=  \mathbb{E} \mathcal{L}(D, \sigma (w)),
\] 
for all $w$ in $W$.
In other words, the algorithm $\mathcal{L}$ gives on average  the same rating to $w$ and $\sigma (w)$ when trained on $D$.
\end{thm}
\begin{pf}
We have 
\[
\mathbb{E} \mathcal{L}(D,w) = \mathbb{E} \mathcal{L}(\sigma (D),\sigma (w)) =  \mathbb{E} \mathcal{L}(D, \sigma (w))
\]
where the first equality follows from the invariance of $\mathcal{L}$ under $\sigma$, and the second inequality follows from the invariance of $D$ under $\sigma$.
\end{pf}

\section{Experiments}

We demonstrate the consequences of our theorems in a computational experiment where we use a deep neural network to 
learn the grammar described in our introduction. The networks are trained using data in which two-letter words with two identical letters are good, and two-letter words with two different letters are bad. The network is then asked to assess novel words containing segments it has not seen in the training set.
Randomness enters into the training of these networks in various places and so Theorem~\ref{thm:randomResult} is the relevant result in this case. Consequently, we do not compare individual trainings of the network on the novel stimuli. For each novel stimulus we train the network numerous times and take the average score over all the trainings. It is these scores that are compared between stimuli.

\subsection{Task and Dataset}

Before discussing the neural network learners that were tested, we describe the
dataset and task that was required of them. As before, our set of words $W$ consisted of 
all two letter words with letters running from A to Z.
The training set consisted of the 24 words AA, BB, $\ldots$, XX paired with rating 1, along with 48 randomly generated words with mismatched segments taken from the list A, \ldots, X, each paired with rating 0.

To assess the ability of the learner to generalize to novel inputs, after training we tested it on the words
\[
\mbox{YY}, \ \ \mbox{ZZ}, \ \ \mathbb{X}\mbox{Y}, \ \ \mbox{YZ}, \ \ \mathbb{X}\mbox{Z}, \ \ \mbox{ZY},
\]
where $\mathbb{X} \in \{ \mbox{A}, \mbox{B}, \dots, \mbox{X}\}$ were randomly selected.
For each learner, the experiment was independently repeated 40 times
with different random seeds.

\paragraph{Encodings.}
We distinguish two different representations for the segments A to Z, namely
the localist and distributed encodings. Both of these representations use a fixed
length bit string. However, while localist codes (also known as 1-of-$k$
encoding) are constrained to include a single non-zero bit, distributed codes
can be any arbitrary combination of $k$ bits, for some fixed $k$.
Distributed encodings are a much more compact representation of data;
indeed, for the same string-length $k$, we can represent an exponentially large
number of segments $2^k$.
The experiment was run on both types of encoding with $k=26$.
When distributed encoding was used, codes for each letter were randomly generated each repetition,
so the exact encoding of the segment X, for instance, is almost
certainly different between two repetitions of a given run.

\subsection{Neural Network Learners}

We tested our theoretical findings on the most popular model in the machine
learning literature today: the artificial neural network.
The words were fed into the neural network by simply concatenating
the two 26-bit codes of their letters.
We experimented with many different architectures, ranging from one to three
hidden layers, and from 256 to 1024 units per layer, with $\tanh$ nonlinearities
for all hidden units.
We trained the models via backpropagation using an iterative quasi-Newton
gradient descent algorithm called the limited memory
Broyden-Fletcher-Goldfarb-Shanno method (L-BFGS), with a maximum of 100
iterations.
Both the neural network and its optimization are implemented in \texttt{torch} \cite{torch}.

\subsection{Results}

\begin{figure}[H]
\begin{center}
 \includegraphics[width=8cm]{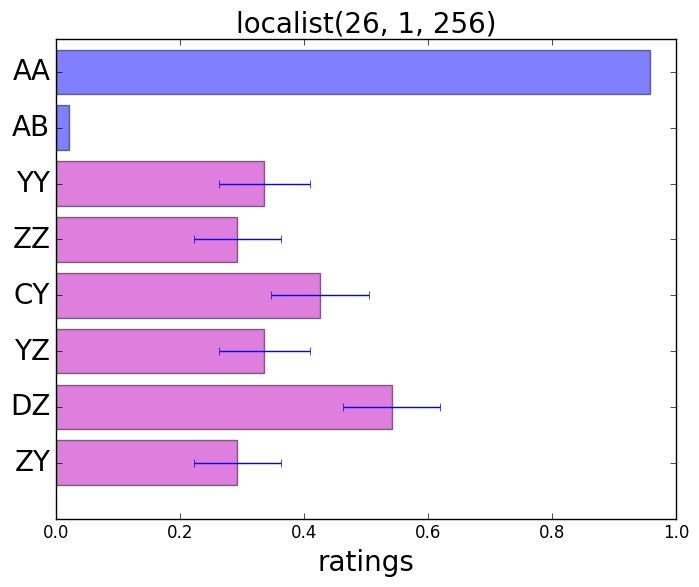}
  \includegraphics[width=8cm]{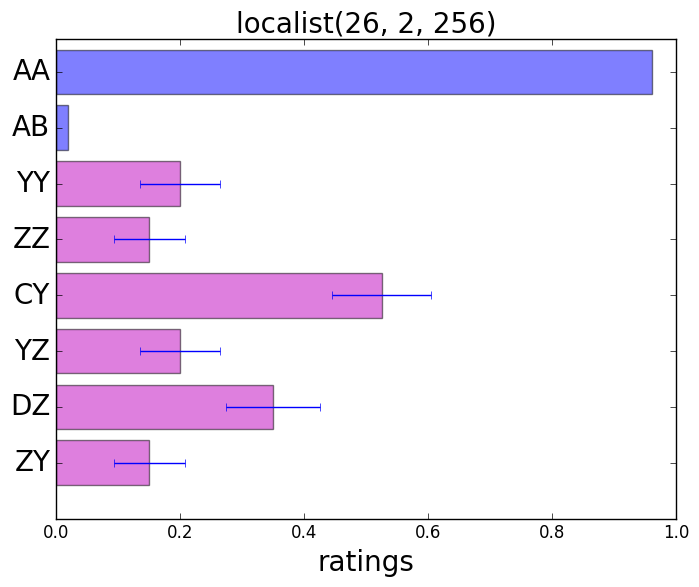}
   \includegraphics[width=8cm]{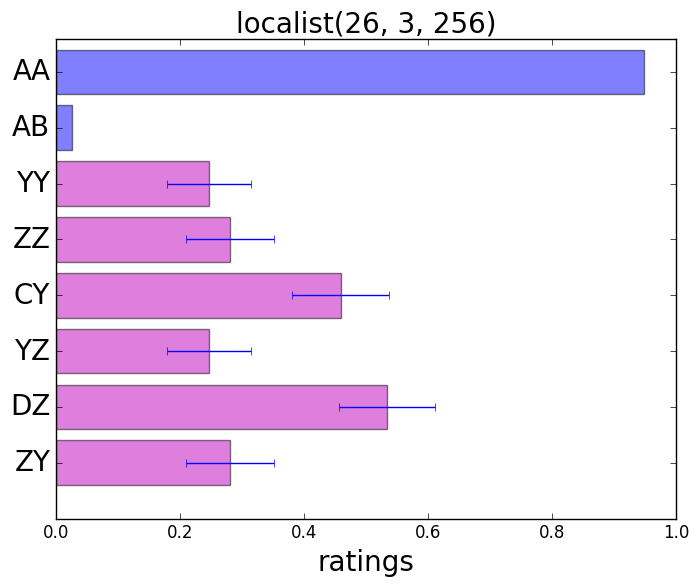}
\end{center}
\caption{Scores for various words for the network with localist encoding for 1, 2, and 3 hidden layers. } 
\label{fig:resultsLocalist}
\end{figure}

We present results for the case of each hidden layer having 256 units, as the results are similar for more units per hidden layer.
In Figure~\ref{fig:resultsLocalist}, for the localist encoding, we plot the
average 
score output by the neural network for each of the test words
above, for 1, 2, and 3 hidden layers.
In addition, the averaged training scores are reported in the top two bars of
each panel.


\begin{figure}[H]
\begin{center}
 \includegraphics[width=8cm]{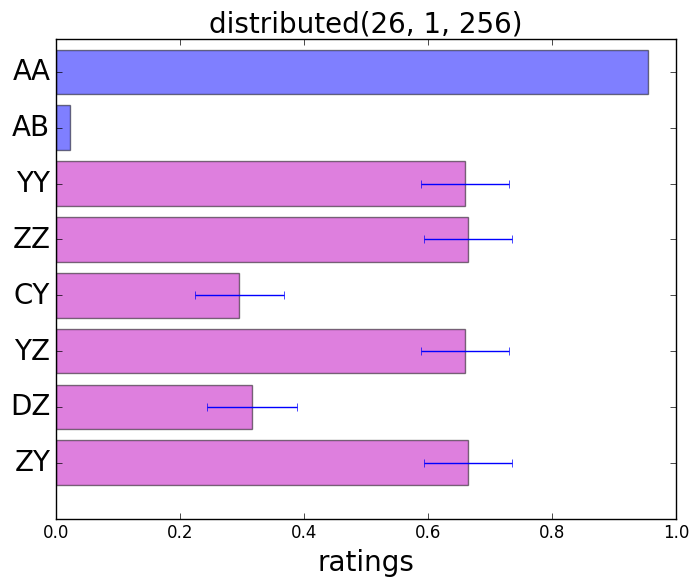}
  \includegraphics[width=8cm]{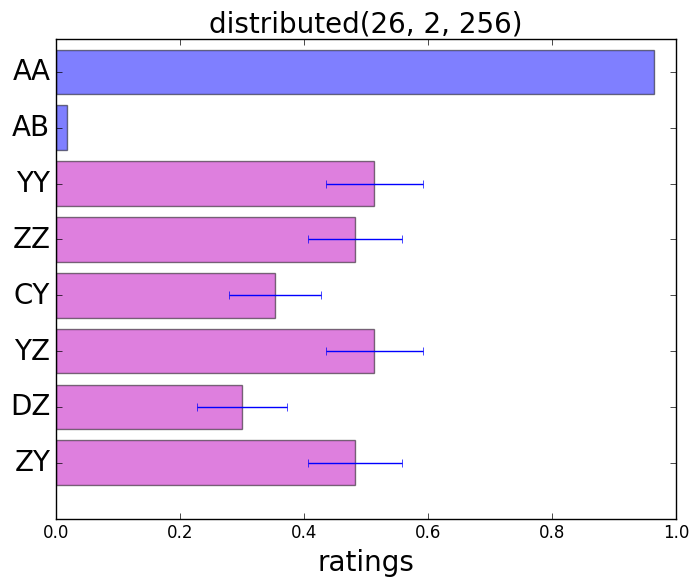}
   \includegraphics[width=8cm]{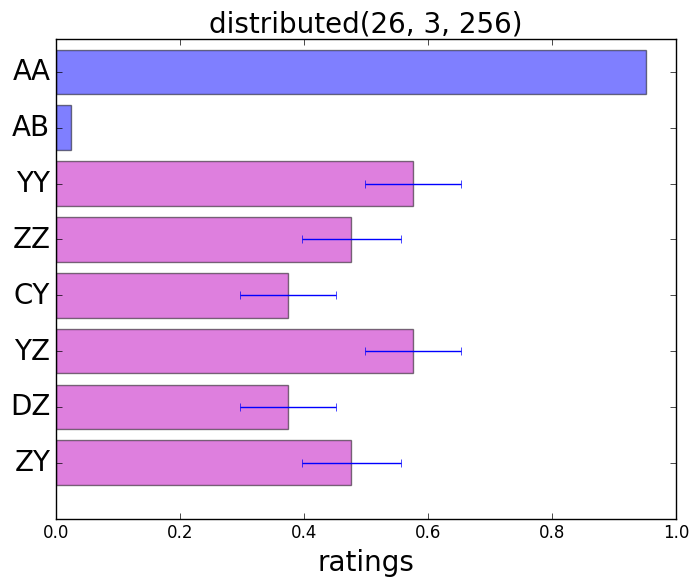}
\end{center}
\caption{Scores for various words for the network with distributed encoding for 1, 2, and 3 hidden layers. } 
\label{fig:resultsDistributed}
\end{figure}

\newpage

Looking at the top plot in the figure, showing the results for one hidden layer, the words YY and ZZ get scores of around $0.3$ in contrast to the score of near $1$ given for the well-formed input AA. The networks are unable to determine that YY and ZZ are grammatical. Likewise, the other test words with differing segments and containing the segments Y or Z have scores ranging from approximately $0.3$ to $0.5$. The networks are not able to distinguish between grammatical and ungrammatical words in this case.

The ability of the networks to generalize to novel inputs is not improved by adding further hidden layers. The second and third plots in Figure~\ref{fig:resultsLocalist}, corresponding to two and three hidden layers, show very similar results to the first.  To within statistical accuracy, the scores for YY, ZZ, YZ, and ZY are all the same. The networks are not able to discriminate between grammatical and ungrammatical words when the words included the novel segments Y and Z.

This poor performance  is perhaps not surprising for the localist encoding, as observed by Marcus \cite{marcusbook}: 
    in the localist encoding, introducing new segments correspond to activating
    new input units that were never active during training, and therefore whose weights never
    changed from their random initializations.
 However, in Figure~\ref{fig:resultsDistributed} we show that the poor performance remains
 true in the case of distributed representations.
 In the first plot,  we show the results for a single hidden layer. 
The networks give a rating higher than 0.5 for both YY and ZZ, which is higher than the score given by the localist networks, but the same high rating is given to the words YZ and ZY. A similar pattern is repeated for the two and three-layer case. The networks are not able to discriminate between grammatical and ungrammatical words containing novel segments, even when distributed representations are used.

\section{Discussion}

That connectionist networks are unable to generalize what are sometimes called ``algebraic" rules to novel inputs is not a new observation \cite{marcusbook,berentbook}. Our contribution has been to give a formalized description and proof of this phenomenon. Furthermore, our results and computer experiments reinforce that Deep Learning, in the form of the ability to  train connectionist networks with multiple hidden layers, does not alone overcome these limitations.


\section{Acknowledgments}

The authors thank Nilima Nigam for comments on an earlier draft of this manuscript.
PT was supported by an NSERC Discovery Grant, a Research Accelerator Supplement, and held a Tier II Canada Research Chair. BS was supported by an NSERC Discovery Grant.

\vspace{1cm}

\bibliographystyle{apacite}

\setlength{\bibleftmargin}{.125in}
\setlength{\bibindent}{-\bibleftmargin}

\bibliography{CogSci_Template}

\end{document}